\documentclass[11pt]{article}
\usepackage[margin=1.05in]{geometry}
\usepackage[T1]{fontenc}
\usepackage{graphicx}
\usepackage{booktabs}
\usepackage{array}
\usepackage{microtype}
\usepackage[numbers,sort&compress]{natbib}
\usepackage{xcolor}
\usepackage{hyperref}
\hypersetup{colorlinks=true,linkcolor=blue!55!black,citecolor=blue!55!black,urlcolor=blue!55!black}

\title{Parsing the Stream: A Live Trace Model for Long-Horizon Agents and Their Observers}
\author{Egor Pakhomov\\Salesforce AI Research\\\texttt{epakhomov@salesforce.com}
\and
Erik Nijkamp\\Salesforce AI Research\\\texttt{erik.nijkamp@salesforce.com}}
\date{}
\begin{document}
\maketitle
\begin{abstract}
A long-horizon agent's trace outgrows both of its consumers: the human observer monitoring the run, and the agent itself, whose bounded context the trace must be folded back into. We present a live trace model, an append-only event ledger folded incrementally into typed run state and compiled into per-consumer views, and evaluate it for both consumers against deterministic ground truth. For the observer side, evaluated with an LLM reader as proxy, the compiled view answers monitoring questions using approximately 14$\times$ and 15$\times$ fewer input tokens (by reader) and at 5--7$\times$ lower cost than a budget-capped single-call reading of the raw trace, with higher accuracy (0.85--0.87 versus 0.48). Because the questions were co-designed with the view schema, we treat the token and cost reduction, conditional on schema coverage, as the transferable result. For the agent, on 120-link sequential-dependency tasks, mechanisms that maintain the task's running statistic in per-step state succeed where full-context prompting fails (30/30 versus 8/30 under a clean protocol, $n=30$, labeled descriptive owing to benchmark--system co-development); a prompt-level scratchpad matches the fold's accuracy at lower cost, and a two-arm decomposition attributes the fold's accuracy to its deterministic aggregate and its cost advantage to its compactness. The fold's remaining value over cheaper alternatives is deterministic auditability and serving the observer from the same state. We derive eleven candidate requirements for trace folding from observed failures and delimit them with an order-sensitive task family on which the fold ceases to help. Code, benchmarks, a regenerable synthetic corpus, and all workbench traces are released.
\end{abstract}

\section{Introduction}\label{sec:1}

A long-horizon agent run continuously produces a single artifact: its trace. Two consumers depend on this trace, and both are underserved by its raw form. The human observer needs to determine, mid-run, what the agent is doing, what has settled, and what remains pending; in our corpus, twelve real sessions total 112 MB, and a frontier-tier model reading their raw tails answers basic monitoring questions at 0.479 accuracy while consuming 779K input tokens across the twelve-transcript panel. The agent itself must fold the trace back into a bounded context window; on our sequential-dependency workbench, a full-context worker's cumulative billed input reaches 2.37M tokens at 120 links (per-call context approximately 33K tokens and growing) while its success rate falls to 7/30 under the development-era injected-error schedules (8/30 under the clean protocol of \S\ref{sec:5.2}). The uncached flat arm bills \$7.49 per run, while the cached conversational variant is inexpensive (approximately \$1.06) yet fails entirely (0/10); cost and failure are therefore separate problems, and neither form of full context addresses the latter.

Today these two consumers are served by two separate systems built over the same stream: observability tooling on one side and context management on the other. This paper models the trace once and compiles per-consumer views from a single substrate. The central measured claim is deliberately narrow: on accumulation tasks, a deterministic bounded fold of the trace shows no observed accuracy disadvantage, at the tested sample sizes, against an arm that retains full history alongside the harness-computed aggregate, at roughly one-quarter the cost, and, unlike equally accurate worker-side alternatives, serves a human observer from the same state with an auditable provenance trail.

\begin{figure}[t]\centering
\includegraphics[width=\linewidth]{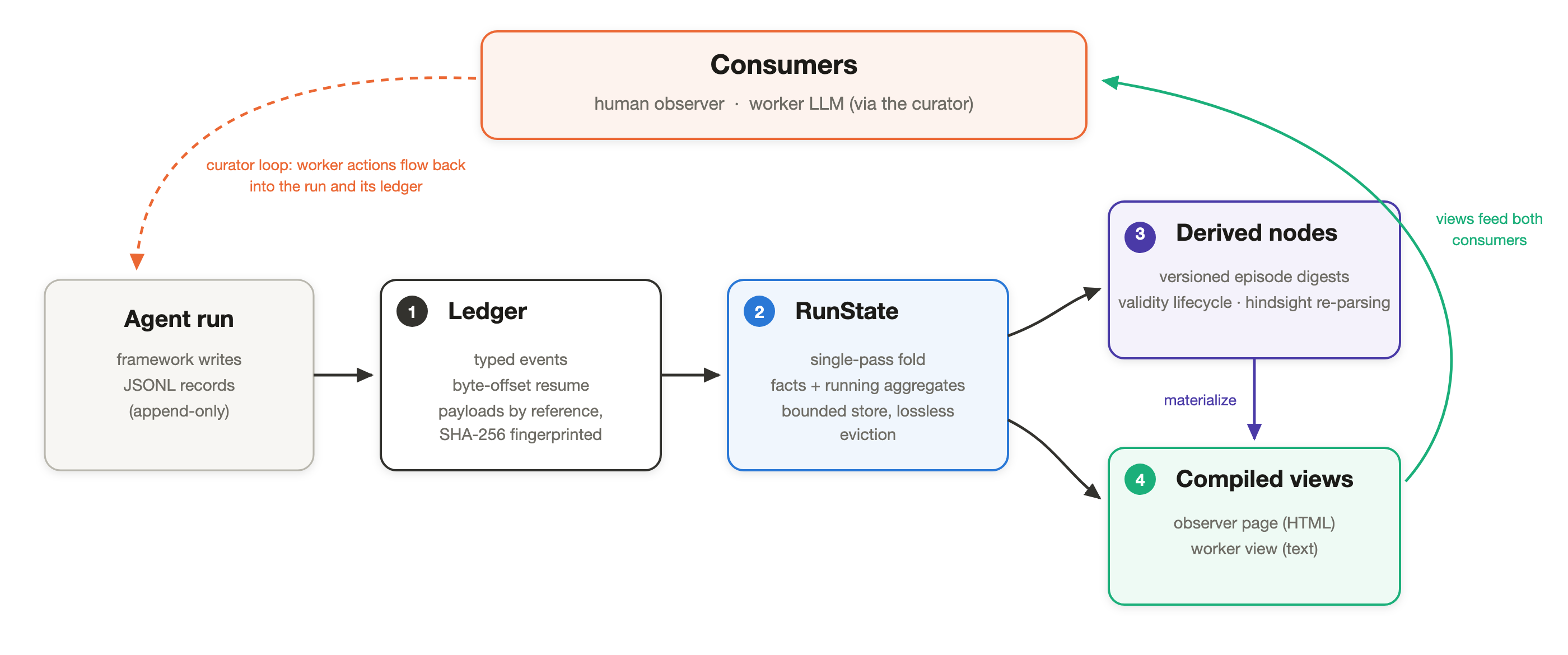}
\caption{The four-layer live trace model. The framework's append-only JSONL stream is parsed into a typed ledger, folded into RunState, materialized into versioned derived nodes, and compiled into per-consumer views; the curator loop closes the cycle --- the worker's actions become part of the run, whose records re-enter the ledger.}\label{fig:architecture}
\end{figure}

The live trace model, implemented in tracelab, is a four-layer stack (Figure~\ref{fig:architecture}): an append-only ledger of typed events with byte-offset resume; a single-pass fold --- the incremental reduction of the event stream into typed state --- producing RunState (tools, files, turns, costs, and facts with source-scoped identity and deterministic aggregates); versioned derived nodes with an explicit validity lifecycle and hindsight re-parsing; and per-consumer compiled views, an observer page and a compact worker view, both compiled from the same state. A curator loop refreshes the worker's view from the worker's own recorded trace, so the system observes its own execution through the substrate it serves.

We make the following contributions, each grounded in the measurements that support it:

\begin{enumerate}
\item \textbf{The architecture}: an incremental, resumable, cache-aware parse of an agent's stream, with design rules that each trace back to a measured failure (per-message usage deduplication that prevents up to 3.49$\times$ token-accounting inflation, terminal validity states that cannot be reopened, and hindsight re-parsing).
\item \textbf{Eleven requirements on the fold}, each surfaced by a live failure during a sequential development ladder and pinned by a regression test or confirmation rerun, together with their measured boundary: an order-sensitive task family on which the fold's aggregates do not apply and it ceases to help (\S\ref{sec:5.3}).
\item \textbf{COMPREHEND}, a live-run comprehension benchmark with mechanically generated questions and deterministic grading, whose corpus is fully regenerable from code: a seeded generator rebuilds twelve realistic sessions byte-identically, making the instrument fully auditable without releasing any real data.
\item \textbf{CONTINUE}, a workbench isolating context policy under matched adversity (pre-committed error schedules, retry/repeat separation, costed protocol failures), with a family of controls comprising a scratchpad, a calculator tool, retrieval, observation masking in the style of~\cite{lindenbauer2025complexity}, summarization in two configurations, and a masked-history-plus-notes hybrid (10/10 at 120 links).
\item \textbf{Economics with real accounting}: prompt-cache read/write measurements showing that structure, not intent, determines caching behavior, and the parser's own inference budget treated as an experimental variable, with extractor availability as a first-class axis.
\end{enumerate}

Section~\ref{sec:2} situates the work; \S\ref{sec:3}--\ref{sec:4} describe the model and instruments; \S\ref{sec:5} reports results; \S\ref{sec:6} discusses; \S\ref{sec:7} states limitations; \S\ref{sec:8} lists artifacts. Appendix~\ref{app:A} carries the development history; Appendix~\ref{app:B}, protocol detail; Appendix~\ref{app:C}, the full cell inventory; Appendix~\ref{app:D}, cache accounting; Appendix~\ref{app:E}, COMPREHEND question detail.

\section{Related Work}\label{sec:2}

We organize prior work by the density of existing research.

\textbf{Crowded areas.} Context compression and folding is an active family~\cite{hu2024hiagent,contextfolding2025,resum2025,acon2025,agentfold2025,lu2026longseeker}: pruned histories outperform full history (18.0\% vs 15.0\% on SWE-bench Lite~\cite{sweagent2024}); sharded incremental disclosure degrades multi-turn performance by 39\% while single-turn concatenation retains 95.1\%, isolating presentation as the failure source~\cite{laban2025lost}; and monolithic context rewriting has a documented collapse mode, with~\cite{zhang2026ace} reporting its Dynamic-Cheatsheet baseline shrinking a context from 18,282 to 122 tokens in a single step, motivating ACE's anchored incremental updates, a pattern our derived-node layer adopts~\cite{factory2025compression}. Retrieval-augmented and virtual-context agent memory~\cite{chhikara2025mem0,packer2023memgpt} includes bi-temporal graphs with provenance and validity windows~\cite{rasmussen2025zep}; these systems are established for cross-session memory but unproven as the model of a single live run; the vendors' own papers report conflicting relative effect sizes~\cite{rasmussen2025zep,chhikara2025mem0}, while independently measured compaction effects are small~\cite{sweagent2024,openhands2025sdk}. Agent observability tooling is extensive at the transport layer, with OTel GenAI conventions, span dashboards, and trace stores~\cite{otelgenai,traceplatforms}, but it aggregates spans rather than maintaining a semantic model of what a run is doing. KV-cache economics is well established: append-only prefixes and careful breakpoints are standard practice because mutating a cached prefix forfeits approximately 10$\times$ cached-input pricing~\cite{ji2025manus}, corroborated by a condenser measurement in the OpenHands repository that resolved fewer SWE-bench instances at higher dollar cost through lower cache utilization~\cite{openhands2025sdk} and by systematic caching evaluations across more than 500 agent sessions~\cite{lumer2026cache}.

\textbf{Adjacent foundations.} The layer stack applies event sourcing with CQRS~\cite{fowler2005es} to agent runs, with incremental materialized views as the streaming analogue~\cite{budiu2023dbsp}. Two distinct findings motivate validating anything fed back to a worker: context growth alone degrades success~\cite{zeng2026loca,hong2025contextrot} (Claude Opus 4.5 falls from 96.0\% to 14.7\% as environment descriptions grow from 8K to 256K~\cite{zeng2026loca}), and compaction errors, once introduced, surface within three steps of a paired continuation in 88--100\% of cases~\cite{chen2026slipstream}.

\textbf{Closest neighbors.} VISTA~\cite{xu2026vista} gives the agent a typed, addressable model of its own working memory, but its dashboard is proprioceptive rather than a human observer view. ESAA~\cite{santos2026esaa} applies strict event sourcing to agents, with schema-validated intentions deterministically projected into a hash-verified read model, but is organized around project governance rather than live observability. PROJECTMEM~\cite{malo2026projectmem} deterministically projects a typed event log into an agent-readable summary at cross-session granularity. LangGraph's streaming architecture~\cite{langgraph2026streams} establishes the pattern of a single typed event log driving many live views for frontends, without tying projections to the model's next context. Live trace visualization also exists with the monitor deliberately separated from the agent's context, validated with an $n=5$ expert study~\cite{gao2026got}. These components exist separately; the single-fold-serving-both-consumers architecture measured here targets that remaining intersection. Post-hoc failure taxonomies improve human agreement ($\kappa$ reaching 0.88~\cite{cemri2025mast}) but operate after the run, and trace-understanding evaluations rely on LLM judges or annotation, with the best model at $\sim$11\% on TRAIL~\cite{deshpande2025trail}. We position this work within these seams and claim no priority in them.

\section{The Live Trace Model}\label{sec:3}

\textbf{Ledger.} The source framework writes predominantly one JSONL record per content block; the adapter emits one typed event per block, preserving causal chains, tool-call/result correlation, and cache-aware usage fields, with payloads stored by reference; content-bearing payload records (messages, thinking, tool calls and results, compaction summaries) carry a \mbox{SHA-256} fingerprint of the full original text, so referenced content is tamper-evident even though only a bounded excerpt is inlined; auxiliary records (attachments, session metadata, unrecognized kinds) fingerprint a bounded excerpt. Malformed lines are quarantined and counted; unknown record types are preserved. Under the observed writer discipline (append-at-tail, with no rotation, truncation, or file replacement), a byte offset is a sufficient resume token (violations of that discipline would invalidate it, and handling them is deployment work not covered here): the parser attaches to a live file and never consumes a partial line, and property tests confirm that chunked ingest equals whole-file parsing. Two measured properties are consequential downstream. First, the per-block format repeats message-level usage on every record; without deduplication by API message id, token accounting inflates by up to 3.49$\times$ (the naive-to-deduplicated ratio of accounted input tokens on the session used for verification; corpus-wide inflation ranged roughly 2--3.5$\times$, and the cost ratio tracks the token ratio), a hazard for any consumer of block-exploded transcripts. Second, ingest is inexpensive relative to all downstream processing: 104 MB across the ten largest transcripts parses into 16,737 events in 0.4 s on one Apple silicon performance core (a single warm-cache, single-threaded timing, indicative rather than a benchmark; it excludes optional LLM extraction and view materialization).

\textbf{RunState.} A single-pass fold produces execution position (goal, frontier, pending calls), counters not exposed by the SDK, files touched, and \emph{facts}: values extracted from tool results, keyed by source (\texttt{file:key}), with occurrence identity (repeated observations of the same key each count, rather than newest-wins), a bounded store, aggregate-preserving eviction (evicted numerics fold into per-key running counts and sums, preserving exactly the tracked statistics and nothing else), and re-read idempotence across eviction. The fold is chunking-invariant by property test, and an independent from-scratch recount over raw JSONL reproduces every per-key fact-plus-aggregate sum on five 120-link traces with zero mismatches (Appendix~\ref{app:B}).

\textbf{Derived nodes.} Versioned artifacts (per-turn episode digests) carry a validity lifecycle (current, suspected-stale, invalidated/superseded) with a terminal-state guard and version-sensitive staleness propagation. Hindsight re-parsing makes versioning necessary: an interrupted turn is only recognizable when the next user turn arrives with calls dangling, and a late result must attach to the episode that issued it. The parser revises its account of the past by superseding prior versions rather than rewriting them.

\textbf{Views and the curator.} The observer view is an HTML page: goal, live frontier, anomaly badges, stat cards, episode drill-downs with provenance links, and an honesty watermark stating the materialization point and malformed counts. The worker view is a compact text block (GOAL / NOW / ANOMALY badges when present / COUNTERS / FILES / KEY FACTS, grouped by key with running aggregates). Both compile from one state. The curator re-materializes the worker view from the worker's own recorded trace every $K$ steps ($K=5$ as deployed; \S\ref{sec:5.4} varies the cadence). The following excerpt is the shipped renderer's output at the final step of a 120-link run (abridged; the aggregate line shows requirements 5--7 and the coverage stamp of requirement 11 --- \S\ref{sec:5.3} --- operating together):

\begin{quote}\begin{scriptsize}\begin{verbatim}
# Run model - session workbenc          <- id truncated by the renderer
GOAL: Follow the chain: ... Sum ALL deltas and write 'total = <sum>' ...
NOW: calling done()
ANOMALY[warn] tool_flood: read_file called 136x (cap 100)
COUNTERS: turns=1 tool_calls=138 errors=8 est_cost=$1.71
FILES TOUCHED: node/01762741.txt, ... (122 files)
KEY FACTS (120 in view of 180 extracted -- grouped by key; evicted values fold into [aggregate] lines):
  node/0492c4f5.txt:delta = 32
  ...
  [aggregate] delta: 120 values total (60 folded out of view), sum = 5281 -- ALREADY
  INCLUDES every delta above and every folded one, through node/d3e89d32.txt:delta
\end{verbatim}\end{scriptsize}\end{quote}

The renderer used in the development-era grids contained two header defects: a ``complete list'' label over a partially evicted store, and an extraction count that reported only the in-view total. Both violated requirement 3 (no silent truncation) one level above the fact store they described, and the related coverage ambiguity in the aggregate line produced the failure mode analyzed in \S\ref{sec:5.2}; the header wording and the coverage stamp above are the corrections.

One declared exception to layer determinism: the optional semantic extractor (\S\ref{sec:5.5}) is an LLM. Its outputs are memoized per event, so refolds within a run are reproducible, but not across cache loss or model retirement. ``Deterministic'' in this paper refers to the non-extractor path; the extractor path is validated and provenance-carrying (requirement 10), not deterministic.

\section{Instruments}\label{sec:4}

Every score in this paper is computed against deterministic ground truth, a deliberate contrast with LLM-judged agent evaluation; measured agent outcomes and costs additionally vary widely across scaffolds~\cite{kapoor2025hal}. Ground truth is either constructed (synthetic pathologies, pre-committed schedules), computed by an independent oracle, or mechanically derived from the ledger. No LLM judges are used anywhere: LLM readers are themselves unreliable judges of traces, with the best model scoring roughly 11\% overall on TRAIL~\cite{deshpande2025trail}. The COMPREHEND reader is not a judge; it is the measured consumer of the representation under test, and its answers are graded deterministically. The TRAIL result argues against LLMs assigning scores, not against LLMs being the object of measurement.

\textbf{DETECT} (floor check). Rule-based detectors applied to synthetic traces carrying 48 labeled pathology instances across 30 traces achieve $P=1.0$/$R=1.0$ with zero false positives on clean traces. This establishes implementation consistency rather than real-world validity: on real workloads, a tool-flood threshold fires on legitimately tool-heavy runs.

\textbf{FIDELITY} (floor check). The incremental fold matches a mutation-tested independent oracle on 8/8 bookkeeping fields, and the extended fact/aggregate recount above covers the machinery that carries the agent-side results.

\textbf{COMPREHEND} (observer leg, LLM reader as proxy). A question builder generates six monitoring question types deterministically from ledger ground truth. Readers answer from a capped raw tail, a capped flat log, or the compiled view, and grading is exact-match, substring, or set-F1 (set-F1-or-none for the dangling-calls question). Deterministic derivation does not make ledger-derived ground truth infallible: tool-call/result matching links 3669/3671 pairs = 99.95\% on the real corpus, where the two orphans are end-of-file danglings and are themselves informative. For this reason, the synthetic corpus additionally computes its ground truth independently of the adapter, yielding zero mismatches on turns, top tool, files, and latest ask; the dangling and last-error fields inherit adapter correlation. The scope of the instrument is limited: the questions target fields the view renders by name, so the benchmark measures schema-scoped extraction and retention under a fixed reading budget and cannot distinguish whether the view preserves what observers need from whether it merely contains what the benchmark asks. A deterministic responder would score approximately 1.0 at no additional LLM-inference cost, and that responder is the system itself, since the observer page renders these fields directly from RunState. The LLM-reader instrument instead measures what a consumer without API access extracts from each textual rendering under a budget; it measures the cost of consuming each rendering rather than establishing that an LLM is necessary. Independently specified questions, a full-access tool-using reader, and human subjects remain future work.

\textbf{CONTINUE} (agent leg). A live worker operates on a deterministic workbench comprising the scatter, fix, chain, prose-chain, and altchain task families, with success verified against final environment state. Arms share the worker model, tasks, seeds, and pre-committed per-call-index error schedules; all calls run with the endpoint's extended-thinking mode disabled (Appendix~\ref{app:B}), and whether enabled thinking changes the crossover is untested. These schedules are identical by call index rather than by semantic operation, so arms that make more calls face more injected errors and arms that make fewer face fewer; exposure is therefore endogenous to the policy. Where exposure differs most, in the calculator arm whose two calls per link roughly double it, the direction runs against the successful arm, which nonetheless scores 10/10; for the headline pair, mean per-run call counts are close but not identical (curated $\approx$133, range 128--142; full $\approx$142, range 133--145, reaching the step cap when lost), so the failing arm absorbs modestly more injected errors, a residual confound whose sign we cannot fully determine. A no-injection replication provides the cleaner comparison, first at pilot scale (curated 5/5, full 0/5) and then adequately powered ($n=30$ in both arms with the shipped renderer: curated 30/30, full 8/30; \S\ref{sec:5.2}); the full-context arm's observed success counts under the injected and clean protocols differ by only one (7/30 versus 8/30), indicating that injected adversity is not the dominant cause of the crossover. Retries are distinguished from repeats, and protocol failures are recorded and costed. Under the curated arm's protocol, the worker's context consists of the compiled view plus the last five raw steps, re-materialized from its own recorded trace every five steps, and nothing else. One asymmetry must be stated explicitly: the curated arm's treatment includes deterministic computation over the trace, so the comparison is agent plus trace model versus agent alone. Section~\ref{sec:5.2} accordingly includes a family of state-discipline controls and a two-arm decomposition separating the computation from the boundedness.

\section{Results}\label{sec:5}

\subsection{The observer leg}\label{sec:5.1}

We evaluated 12 real transcripts (112 MB; 70 questions per condition; one reader call per transcript-condition; scores are unweighted means over per-question scores in [0,1]):

\begin{table}[htbp]\centering\footnotesize
\caption{COMPREHEND on the real corpus: accuracy (mean over questions) $\cdot$ input tokens and cost as panel totals over the 12 reader calls, by condition and reader. Questions are co-designed with the view schema (\S\ref{sec:5.1}).}\label{tab:comprehend}
\begin{tabular}{llll}
\toprule
reader & raw tail & flat log & compiled view \\
\midrule
Sonnet 5 & 0.479 $\cdot$ 779K $\cdot$ \$2.37 & 0.621 $\cdot$ 479K $\cdot$ \$1.59 & \textbf{0.871 $\cdot$ 57K $\cdot$ \$0.42} \\
Haiku 4.5 & 0.476 $\cdot$ 652K $\cdot$ \$0.53 & 0.522 $\cdot$ 372K $\cdot$ \$0.32 & \textbf{0.850 $\cdot$ 43K $\cdot$ \$0.08} \\
\bottomrule
\end{tabular}
\end{table}

\begin{figure}[t]\centering
\includegraphics[width=\linewidth]{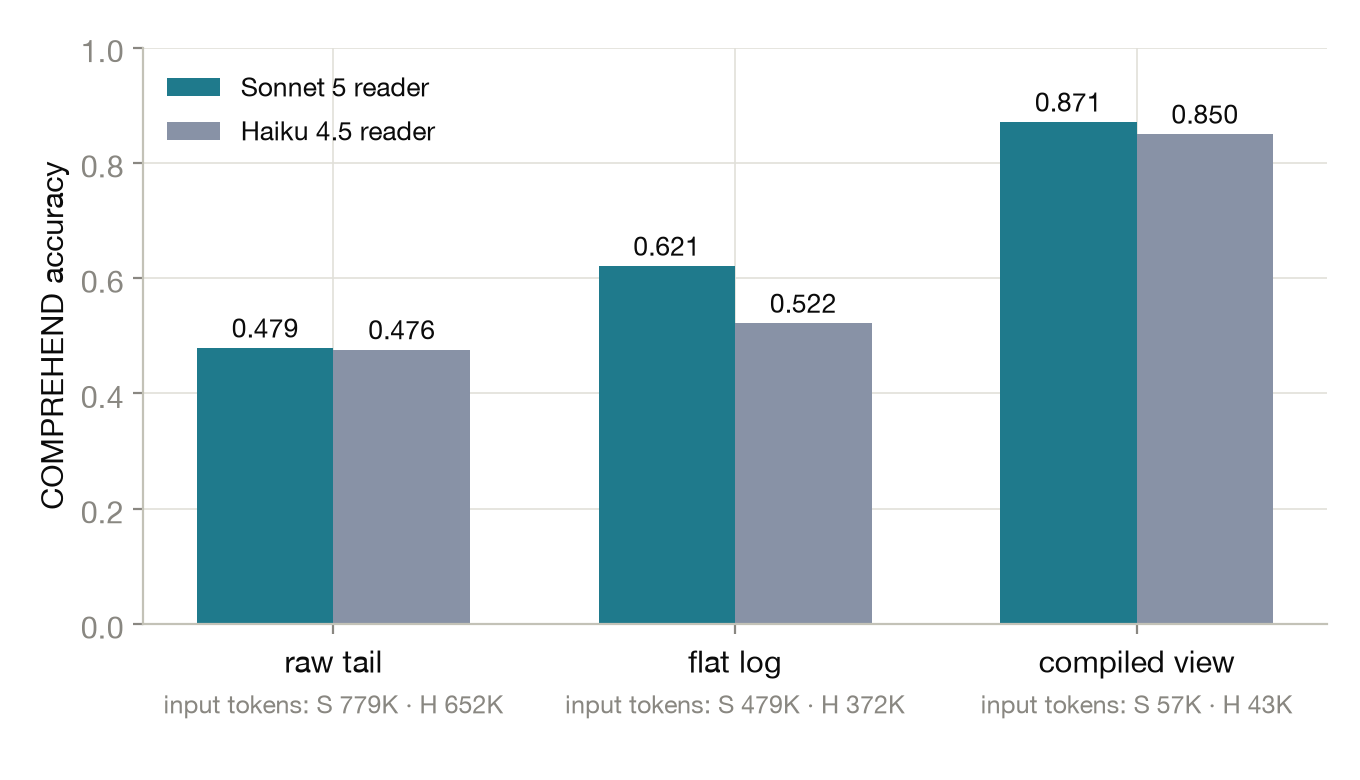}
\caption{COMPREHEND accuracy by condition and reader (12 transcripts; one reader call per transcript-condition). Gray annotations give total input tokens across the panel of 12 reader calls (S = Sonnet 5, H = Haiku 4.5). The view-vs-raw gap is $\approx$0.37--0.39 for both readers; the reader-vs-reader gap within any condition is $\le$0.10.}\label{fig:comprehend}
\end{figure}

Transcript-level bootstrap 95\% confidence intervals from an independent full rerun are: view 0.86 [0.78, 0.93] versus raw 0.51 [0.42, 0.61] for Sonnet 5, and 0.81 [0.74, 0.88] versus 0.46 [0.39, 0.54] for Haiku 4.5; the intervals are disjoint for both readers. These intervals quantify transcript resampling of this fixed, size-selected corpus under one call per cell; they do not capture reader-call noise or any wider population. The rerun's cell-by-cell deltas against the original results span 0.013--0.051 (largest: Haiku flat log). Two calls per cell provide only a first indication of single-call reader variance, not a bound; repeated-reader resampling remains future work.

For queries addressable directly against RunState, a deterministic responder would answer all six fields at approximately 1.0 with no additional LLM-inference cost (\S\ref{sec:4}); Table~\ref{tab:comprehend} reports consumption of the textual renderings. The condition gap ($\approx$0.37--0.39, view versus raw) substantially exceeds the reader gap ($\le$0.10 within any condition; Figure~\ref{fig:comprehend}). Consistent with this ordering, the smaller reader on the compiled view (0.850, \$0.08) outperforms the frontier reader on the raw tail (0.479, \$2.37). The resulting $\approx$30$\times$ cost ratio conflates model-tier pricing with the representation change, so we report it here rather than in the abstract; the same-reader 5--7$\times$ reduction is the appropriate summary figure. The cost multiple is smaller than the token multiple because output-token and per-call costs do not shrink with the input.

The per-question decomposition is more informative than the pooled score. Whole-run aggregation questions collapse without the view (files: 0.13--0.20 raw versus $\sim$1.0 view), while two questions are condition-insensitive for different reasons. The dangling-calls question scores 0.917 in every condition because eleven of twelve transcripts share the majority ``none'' answer; this item measures little and compresses pooled gaps; macro-averaging over the remaining five question types widens view versus raw to 0.850 versus 0.385 for Sonnet 5 and 0.815 versus 0.382 for Haiku 4.5. The latest-ask question scores 0.833 from the raw tail because recency survives truncation, although it declines in the flat log. The view's advantage therefore concentrates precisely where the 100K-character reading budget (57.8K tokens; 1.73 chars/token on escape-dense JSONL) cannot reach: retention through aggregation, which is the function the system exists to provide. An uncapped single-call reader is not available as a baseline: the median transcript is $\sim$11 MB $\approx$ 6M tokens at the measured 1.73 chars/token, beyond the context window of any model in the deployed stack; this infeasibility is the problem the system addresses. A tool-using multi-call reader is the natural future baseline.

The flat log serves as the control for this concern: a compacted whole-run rendering under the same reading cap (receiving 8$\times$ more actual tokens than the view, 479K versus 57K panel-wide) recovers one of the two whole-run aggregation questions only partially (files 0.705 versus 1.0 for Sonnet 5; turn count stays at 0) and still trails the view by 0.25 (Sonnet 5) and 0.33 (Haiku 4.5). The view's advantage over this generic compaction is therefore not attributable to tail truncation alone. Whether a task-aware structured projection other than ours would close the gap is untested, and the token reduction is itself conditional on the schema covering what the consumer asks: a projection can be made arbitrarily cheap by omitting whatever is not tested. On the released code-as-corpus replication (twelve seeded synthetic sessions, 11.4 MB, generator ground truth matching adapter ground truth with zero mismatches on the four generator-checkable fields), the pattern holds: view 1.000/0.944 versus raw tail 0.673/0.600. Raw-tail scores are higher there because sessions are smaller, consistent with the budget mechanism.

\subsection{The agent leg: the crossover and its controls}\label{sec:5.2}

\begin{figure}[t]\centering
\includegraphics[width=\linewidth]{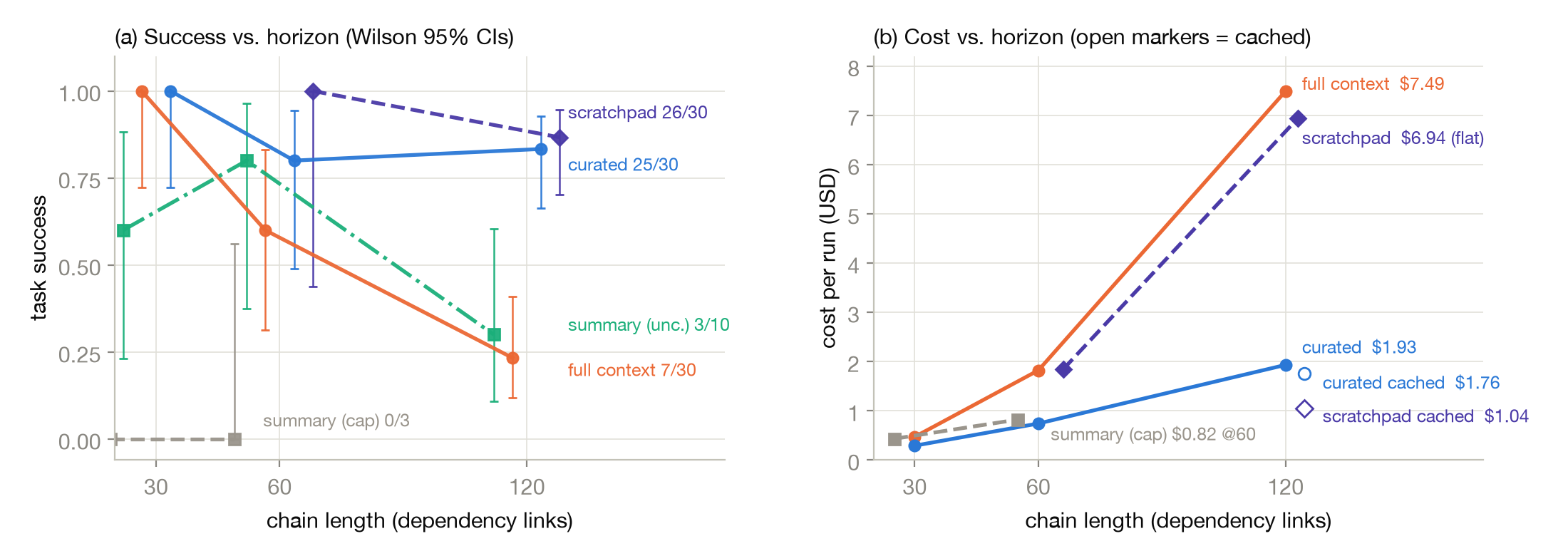}
\caption{The development-era grid (pre-stamp renderer, injected error schedules). (a) Success by horizon with Wilson 95\% intervals (per-arm marginals; cell sizes vary and are listed in Appendix~\ref{app:C} --- denominators appear in each label). (b) Cost per attempted run; solid and dashed lines are uncached, open markers are cached cells. Word-capped summarization was run only to 60 links, and the scratchpad arm was not run at 30; the calculator, conversational-full, and retrieval arms are tabulated in \S\ref{sec:5.2} and Appendix~\ref{app:C} rather than plotted. Table~\ref{tab:primary}, not these 120-link cells, is the primary comparison.}\label{fig:crossover}
\end{figure}

The chain family makes horizon the treatment: each file names a delta and the next file, the answer is the sum, and \texttt{search} is disabled. We report the primary comparison first (Table~\ref{tab:primary}), under the final protocol (shipped renderer, no injected errors, $n=30$ per arm, identical seeds), and then the development-era grid and controls from which the design emerged (Table~\ref{tab:devgrid}; Figure~\ref{fig:crossover}).

\begin{table}[htbp]\centering\footnotesize
\caption{The primary CONTINUE comparison at 120 links under the final protocol (shipped renderer, no injected errors, identical seeds, $n=30$ per arm). Cells are labeled descriptive (\S\ref{sec:5.2}); costs are as-deployed (\S\ref{sec:5.4}).}\label{tab:primary}
\begin{tabular}{>{\raggedright\arraybackslash}p{5.3cm}llll}
\toprule
arm (120 links, final protocol) & n & success & \$/run & cache \\
\midrule
curated view (fold) & 30 & \textbf{30/30} & \$1.59 & cached \\
scratchpad (full context + note-field instruction) & 30 & \textbf{30/30} & \$0.97 & cached \\
full context (flat) & 30 & 8/30 & \$7.13 & uncached \\
\bottomrule
\end{tabular}
\end{table}

Cache eligibility differs by arm: the flat full-context prompt does not cache, and \S\ref{sec:5.4} shows caching can reorder cost rankings where the conversational form was tested, so the cost column compares configurations as deployed, not cache-matched ones. The success column is unaffected.

Paired against full context on shared seeds, the curated arm has 22 sole successes and 0 sole failures (exact McNemar, two-sided, $p\approx 5\times10^{-7}$, descriptive under the adaptive-design caveats below). The fold and the scratchpad are indistinguishable on success under this protocol; the fold costs about 1.6$\times$ as much and provides deterministic, auditable state and the observer view from the same fold. Development-era cells at 120 links were extended to $n=30$. The extension was declared in the run log before launch but after the $n=10$ results were known; it is therefore a declared extension rather than a preregistration. The extension-only cells (16/20 vs 4/20, two-sided Fisher exact $p\approx 3.6\times10^{-4}$) confirm the effect independently of the triggering sample, and excluding the three development seeds (22/27 vs 6/27) does not alter the result.

\begin{table}[htbp]\centering\footnotesize
\caption{The development-era CONTINUE grid by horizon (pre-stamp renderer, injected error schedules); cell sizes vary and denominators appear in each entry.}\label{tab:devgrid}
\begin{tabular}{>{\raggedright\arraybackslash}p{0.7cm}>{\raggedright\arraybackslash}p{3.4cm}>{\raggedright\arraybackslash}p{3.0cm}>{\raggedright\arraybackslash}p{6.9cm}}
\toprule
links & full context & curated view & other arms \\
\midrule
30 & 10/10 $\cdot$ \$0.48 & 10/10 $\cdot$ \$0.29 & summary (capped) 0/3 $\cdot$ summary (uncapped) 3/5 \\
60 & 6/10 $\cdot$ $\sim$\$1.82 ($n=10$ flat grid; the $n=3$ cache-study cell scored 0/3 at \$2.03) & 8/10 $\cdot$ \$0.74 & summary (capped) 0/3 $\cdot$ mask~\cite{lindenbauer2025complexity} 0/3 $\cdot$ summary uncapped 4/5 ($\sim$\$1.06) $\cdot$ scratchpad 3/3 (\$1.84) \\
120 & \textbf{7/30} $\cdot$ \$7.49 & \textbf{25/30} $\cdot$ \$1.93 (cached 3/3 $\cdot$ \$1.76) & conv. full 0/10 (\$1.06 cached) $\cdot$ retrieval 0/10 $\cdot$ summary uncapped 3/10 ($\sim$\$2.84 incl. summarizer) $\cdot$ mask+notes hybrid 10/10 ($\sim$\$2.98) $\cdot$ \textbf{scratchpad cached 26/30 (\$1.04 mean at $n=30$; the $n=3$ pilot averaged \$1.10)} $\cdot$ calculator tool 10/10 (\$14.88) \\
\bottomrule
\end{tabular}
\end{table}

At 120 links, the development-era full-versus-curated separation (7/30 vs 25/30) is the second adequately powered comparison, alongside the primary clean-protocol cell above. Because the arms share seeds, the paired analysis is primary: of 30 shared seeds, 19 are curated-only successes, 1 is full-only (seed 11), 6 are successes in both arms, and 4 in neither (exact McNemar, two-sided, $p\approx 4\times10^{-5}$; the unpaired two-sided Fisher $p\approx 6\times10^{-6}$ is reported for continuity, and the per-seed table is included in the scoreboard). Because the task family, the mechanism, and the decision to extend were all developed adaptively against this generator, we label these p-values descriptive rather than confirmatory; the confirmatory experiment, a frozen system evaluated on externally authored tasks, has not been run.

We regard this family of controls, rather than the single crossover, as the central finding of the section. Every mechanism that carries the running statistic in per-step state succeeds. The curated fold scores 25/30. Prompt-cached self-managed notes score 26/30 at $n=30$ (\$1.04/run); this arm is implemented as full context plus a single instruction to emit a per-step note field, with no harness machinery, so it also serves as a prompting-only control: one instruction moves full context from 7/30 to 26/30, indicating that prompting is itself a first-order treatment. The masked-history-plus-notes hybrid scores 10/10 ($\sim$\$2.98), with observations elided beyond the last five steps and worker notes persisting in the action record; it bounds raw context while retaining worker-authored state. An environment calculator tool scores 10/10 (\$14.88; each link costs two calls). The remaining arms perform far worse: plain full context (7/30), conversational-form full context (0/10; repackaging does not avert the failure, although the flat-versus-conversational spread of 7/30 vs 0/10 indicates that packaging matters even though neither form suffices), retrieval over the worker's own trace (0/10), and word-capped summarization (0/3 at 30 and 60 links; the arm was stopped before 120). The dividing line is not external state as a category, since the retrieval arm stores the entire trace externally and still fails, but whether the mechanism carries the task's needed statistic explicitly in its per-step state, a bounded object even when the surrounding context is not: relevance ranking cannot select ``all of them,'' which is what accumulation requires. This criterion can be checked before running an arm, by asking whether its per-step context carries the current partial result explicitly. It is consistent with every arm in Table~\ref{tab:devgrid}, but as a prediction it has been tested exactly once (the hybrid, built and queued with success expected); we therefore treat it as a hypothesis with one forward confirmation rather than an established regularity.

Failure analysis of the curated arm's five 120-link misses shows that in every case the compiled view displayed the exactly correct total (the fold was correct in 30/30 runs), and the worker then wrote that total plus one already-included delta (+37/+78/+79/+31/+71; in each case the delta of the last file covered by the final view refresh, where fold coverage and the raw last-5 window overlap). The residual failure mode is therefore last-mile ambiguity rather than state corruption: the worker cannot tell whether the rendered aggregate already covers the file it just read. The design implication is a machine-facing analogue of the observer view's honesty watermark: aggregates should state their own coverage. We implemented this after the frozen grids (each aggregate line now ends ``ALREADY INCLUDES every <key> above and every folded one, through <last source>'') and reran the five failing seeds plus five passing controls: 5/5 recovered and 5/5 held. These post-hoc fix cells are reported separately from the frozen 25/30 and are never pooled with it (the frozen headline ran on the pre-stamp renderer, so headline and fix were never simultaneously active); they constitute a regression check on the motivating failures, not independent confirmation. A stronger test, rerunning the shipped (post-stamp) renderer on the full 30-seed grid under the injected schedules, scores \textbf{29/30} (\$1.77/run); the single miss is a new and rarer slip (the worker wrote +4 over a view total that refolds to exactly correct), not the double-add the stamp eliminated. The final protocol of Table~\ref{tab:primary} removes two remaining design confounds (superseded renderer; policy-endogenous error exposure) and is the primary comparison; its cells, like all cells in this section, remain labeled descriptive owing to generator co-development, a caveat no protocol change can remove. The curated numbers thus form a sequence from the development-era grid to the final protocol: 25/30 (frozen pre-stamp renderer, injected errors; the development-era grid whose failure analysis produced requirement 11), 29/30 (shipped renderer under the same injected schedules), and 30/30 (shipped renderer, no injection). None are pooled; each is a separate cell in the scoreboard. Requirement 11 was surfaced by failure analysis, regression-verified on the motivating seeds, and then borne out on the full-grid rerun.

One possible reading of the calculator cell is that CONTINUE measures arithmetic rather than context management. The scratchpad cell separates the two: the scratchpad worker performs the same arithmetic itself, in context, one increment per step, and succeeds 26/30, while the full-context worker, holding every needed number in its window, fails at the same computation (7/30). The bottleneck is not the ability to add but the maintenance of a running result across the horizon, as opposed to recomputing it from 120 scattered values at the end; every disciplined-state mechanism provides this maintenance and undisciplined history does not. Two deconfounding arms make the decomposition explicit. Holding computation equal and varying boundedness, the compiled view plus full raw history scores 10/10 at \$7.63/run. Holding boundedness equal and varying computation, the same bounded view with aggregates stripped and every value kept in sight scores 4/10 at \$2.14/run, its failures being end-stage arithmetic errors of the kind documented in Appendix~\ref{app:A} (two runs wrote partial-prefix sums, scored 0.81 and 0.78). In these $n=10$ post-revision cells (post-stamp renderer in both arms), at this horizon and on this family, the deterministic aggregate carries the accuracy and boundedness carries the cost: the aggregate-carrying bounded view runs $\approx$4$\times$ cheaper than the full-history arm that matches it (deconfound A at \$7.63 versus the post-stamp curated arm at \$1.77). The crossover is thus an arithmetic-maintenance result reachable through either mechanism, and the fold's claim is correspondingly narrower and firmer: it provides accuracy indistinguishable at these sample sizes from full history with the answer present, at roughly a quarter of the price, with an audit trail. We also measured a lower bound: a minimal harness accumulator, a fifteen-line regex tally over the worker's own past observations prepended to the last five raw steps, with no ledger, fold, or view, scores 10/10 at \$0.67/run, the cheapest cell in Table~\ref{tab:inventory}. On this family, that minimal artifact is sufficient. Everything the fold carries beyond it (typed state, provenance, eviction discipline, and the observer page compiled from the same state) is justified by the second consumer and the audit trail rather than by chain accuracy. This delineates task-specific state from a trace model, and the paper's claim rests on the trace-model side of that line.

The tradeoff is as follows. The curated view is not the cheapest disciplined option (cached notes are, at matched n); its distinct value lies in a compact per-call context ($\sim$6K tokens at 120 links versus $\sim$33K and growing), independence from worker note-keeping discipline, deterministic and auditable state rather than unvalidated prose, and serving the observer from the same fold. At short horizons, the negative result stands: on scatter, full context scores 1.000 while structure-only curation scores 0.20, and content pinning recovers the curated arm to 1.000 (Appendix~\ref{app:A}).

\textbf{The summarization correction.} Our initial rolling-summarization baseline, word-capped at approximately 400 words, scored 0/3 at both tested horizons. Removing the cap yields 3/5 and 4/5 at 30 and 60 links; these are directional cells, but no longer categorical failure. The contrast between the two variants indicates that a silently binding length budget converts workable compaction into catastrophic compaction with no warning, which is requirement 3 operating one level up, on the compaction mechanism itself. At 120 links, the $n=10$ cell confirms degradation (3/10, at $\sim$\$2.84/run including the summarizer, whose own cost grows with tracked state). Relative to the fold, the residual gap at 60 links is structural (no determinism, no audit, no bound) rather than a success-rate difference these sample sizes can rank.

\subsection{Eleven requirements, and their boundary}\label{sec:5.3}

The requirements are prescriptive only in a limited sense: each one's violation measurably broke continuation or corrupted state in this setting, and the evidence supports no stronger claim. The violation of requirement 10, for example, produced a phantom value on recovered runs without changing the success rate; it corrupted state rather than causing collapse. The curated arm reached its reported performance through a sequential development ladder: each variant failed in live runs, the failure was traced to a missing property, and the property was implemented and pinned. The requirements are therefore hypotheses generated by adaptive development, not factorial ablations. They are consistent so far with the three truly held-out seeds (3/3 at 60, 2/3 at 120; six trials, five successes; the single held-out miss was the requirement-11 double-add, and the held-out seeds have not been rerun on the stamped renderer) and with the $n=30$ grid, which is not held out because it includes development-era seeds; we consider ``validated'' too strong a characterization. We conjecture that requirements 1--7 apply to any bounded fold serving this task class, that 8--10 apply only when extraction runs through an LLM, and that 11 applies to any fold that renders aggregates over partially evicted state. Appendix~\ref{app:A} gives the full development history; the requirements are summarized in Table~\ref{tab:reqs} below.

\begin{table}[htbp]\centering\footnotesize
\caption{The eleven requirements and the live failures that surfaced them.}\label{tab:reqs}
\begin{tabular}{l>{\raggedright\arraybackslash}p{5.0cm}>{\raggedright\arraybackslash}p{6.4cm}}
\toprule
\# & requirement & discovered by \\
\midrule
1 & carry facts, not references & re-reading loops; 0.20$\rightarrow$1.000 on content pinning \\
2 & occurrence identity, not newest-wins & accumulator collapse to one value \\
3 & never truncate silently & renderer kept last-40 facts; totals silently wrong \\
4 & source-scoped identity & equal values from different files merged \\
5 & deterministic running aggregates & end-stage arithmetic slips over 30 values \\
6 & aggregate-preserving eviction & bounded store dropped early values at 120 links \\
7 & re-read idempotence across eviction & one re-read double-counted (+65) \\
8 & canonical key schemas across extraction batches & one accumulator fragmented into phrasing-dependent keys \\
9 & refusal-tolerant batching & a safety classifier refused batches of individually benign machine text \\
10 & verbatim validation of extracted facts & batch nondeterminism invented a phantom value (+41) \\
11 & aggregates state their own coverage & five perfect-view misses at 120 links; the stamp recovers 5/5 --- rerun-verified on its motivating cases and the full grid (\S\ref{sec:5.2}) \\
\bottomrule
\end{tabular}
\end{table}

\textbf{Boundary of applicability.} On alternating-sign chains, where the $k$-th file's delta enters the total with sign $(-1)^{k+1}$ --- so the answer depends on traversal order and the per-key sum is uninformative by construction --- the curated view scores 3/10 at 60 links against full context's 6/10 (fresh runs in both arms; a directional gap these sample sizes cannot rank, two-sided Fisher p$\approx$0.37), and both arms score 0/10 at 120; these boundary cells ran under the development-era protocol. We state this as a limitation of our own system: the fold helps precisely when its preserved statistics match the computation the task requires, and outside that match it can underperform raw history. Worker-side notes, by contrast, adapt to the operation: the cached scratchpad scores 5/5 at 60 and 9/10 at 120 on the same alternating family, so the limitation is specific to fixed harness-side aggregates rather than to external state in general. This sharpens the underlying tradeoff: worker-managed state is operation-flexible but unvalidated, whereas harness-managed state is deterministic and auditable but only as general as its aggregate vocabulary. A fold with ordered aggregates might recover this task family; we have not tested this, and doing so would constitute a further iteration of the co-evolution loop, so we report the boundary rather than extend the system.

\subsection{Cache economics}\label{sec:5.4}

Cache costs were measured with actual cache-read and cache-creation token accounting (precedent:~\cite{lumer2026cache}). Whether requests hit the cache is determined by prompt structure rather than by the author's intent, including the placement of cache directives: a cache breakpoint on a rebuilt monolithic prefix produced zero cache reads and cost more than no caching (\$2.06 vs \$2.03), because lookup occurs only at the new request's breakpoints. Only the append-only multi-turn form produces cache hits, consistent with practitioner and provider caching guidance~\cite{ji2025manus,anthropic2025caching}. With caching, full context at 60 links drops from \$2.03 to \$0.374 (5.4$\times$, $\sim$501K reads/run), below the curated arm's cached figure of \$0.66, which is dominated by refresh-triggered cache writes (Table~\ref{tab:devgrid} lists the same arm uncached at \$0.74). Full context nonetheless remains the less reliable arm in our cells (6/10 vs 8/10 in the uncached development grid at this horizon, and 1/3 in the cached conversational cell itself; directional sample sizes); cost rankings under caching therefore do not correspond to quality rankings.

The curator's refresh cadence acts as a cost parameter: \$0.66, \$0.61, and \$0.55 at refresh intervals of 5, 10, and 20 steps respectively, with success at 1.000 throughout at chain-60. The anticipated staleness penalty did not materialize in the tested cells; however, success sat at the 1.000 ceiling throughout, which would mask any staleness cost. We therefore interpret the cadence result as one-sided in the tested range rather than as evidence that no tradeoff exists.

\subsection{Inference inside the parser}\label{sec:5.5}

Because prose-chain states facts in natural language, the shipped fixed-pattern extractor cannot parse it by construction; it serves as a designed probe rather than a straw-man baseline. To quantify the ceiling attainable through per-format engineering, we additionally evaluated a tuned deterministic parser. At 60 links ($n=3$ per arm), the full-context arm scores 0.333 at \$2.29; the fixed-pattern parser scores 0.000 at \$0.62; and a tuned pattern, written with access to the three prose templates, scores 1.000 at \$0.63. The tuned pattern incurs zero marginal inference cost; its cost is the engineering labor of reading the templates and writing the pattern. This arm confirms that the templates are trivially parsable once known. The small-model parser scores \textbf{1.000 at \$0.80 per run in total, of which \$0.023 is extraction} (approximately 2.9$\times$ cheaper than full context at the arm level). The frontier-model parser scores 0.667 at \$1.01 total, with a cascade trajectory of 0.000$\rightarrow$0.333$\rightarrow$0.667 before plateauing (Appendix~\ref{app:C}). This pair of arms supports one claim: on this content family, the small model matches the tuned pattern without the per-format engineering. Robustness to unseen formats is untested, as no train/test split over templates exists in this design. Extraction cost scales with the number of observations rather than with context length: the small-model parser achieves 3/3 at 120 links for \$0.046 of extraction.

\begin{figure}[t]\centering
\includegraphics[width=\linewidth]{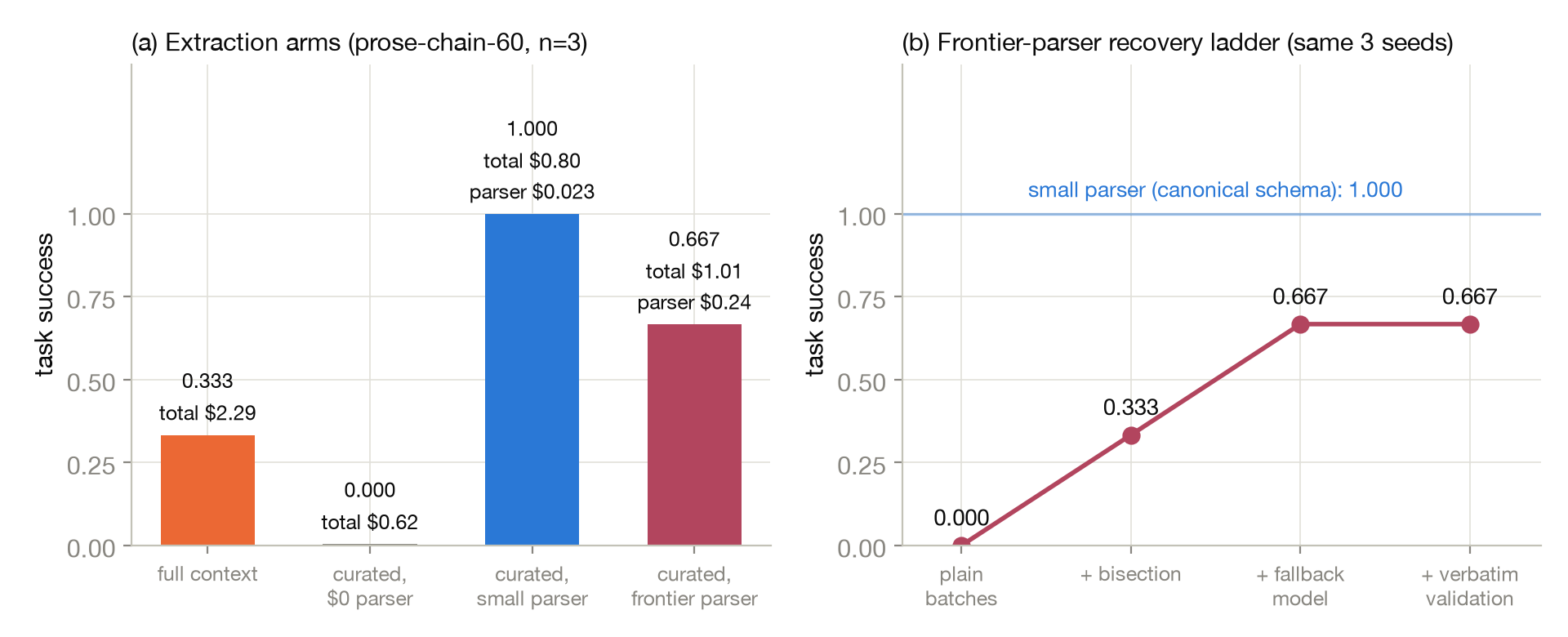}
\caption{Extraction arms on prose-chain ($n=3$ per arm, one content family; extractor availability, model identity, and safety policy are confounded). (a) Success by arm, annotated with total and parser costs (the tuned-pattern arm, 1.000 at \$0.63, is reported in the text and omitted from the plot). (b) The frontier parser's staged recovery on the same three seeds; the horizontal line marks the small parser's 1.000 reference.}\label{fig:parser}
\end{figure}

Accuracy is not monotone in extractor strength, and the mechanism merits explanation (Figure~\ref{fig:parser}). The frontier extractor's safety classifier refused approximately 37 of 85 batched extraction calls on this machine-noise-heavy content (observations resubmitted individually during bisection passed; the batches were flagged as suspicious), whereas the small extractor produced no refusals. Recovering the frontier extractor took requirements 8--10 (canonical key schemas, bisection with a different-model fallback, and verbatim validation), and it still trailed the availability-matched small extractor. In these runs, extractor selection mattered most: the available small extractor outperformed both the stronger refusal-prone model and the engineering spent recovering it. This is an $n=3$ observation on one content family in which model identity and safety policy are confounded, and we do not generalize from it. Availability varies by content and by model, and this variation is not observable until the parser measures it.

\section{Discussion}\label{sec:6}

Three observations recur across the results. First, success on this workbench tracked the requirements rather than the choice of mechanism. The winning fold is deliberately simple; word-capped summarization and plain long contexts fail for reasons the requirements identify; self-managed notes, the masked hybrid, and calculator tools succeed to the extent that they maintain the needed statistic in bounded form; and uncapped summarization holds at mid horizons but degrades (3/10 at 120) as its prose state grows without validation. The requirements serve as the evaluation bar, and CONTINUE functions as a harness in which any candidate mechanism can be added as a policy column.

Second, the COMPREHEND cross-model cell and the CONTINUE crossover reflect a single phenomenon observed from two sides: parsing the stream once, deterministically, moves comprehension work out of every downstream consumer. In the measured cells, a small reader on the compiled view outperforms a frontier reader on the raw tail, and a compact context sustains tasks that a growing one fails, on the task families where the parse matches the task. The parse is paid incrementally, once per event, and is mostly deterministic (with costs on the order of cents when it is not); by contrast, a growing context is re-billed on every call.

Third, the parser is itself a system with its own failure axes, including schema drift, refusals, and batch nondeterminism, which surface only under measurement. A pipeline that batches machine-generated text through safety-filtered models inherits an availability distribution that it has not measured.

The live trace model is one point in a design space, evaluated at modest sample sizes on a single vendor's stack. The elements we hope generalize are the two-consumer framing, the deterministic-ground-truth instruments, the requirement-by-failure methodology, and the boundary result that delimits the scope of the claims.

\section{Limitations}\label{sec:7}

\begin{itemize}
\item \textbf{Sample sizes.} The 120-link headline cells use $n=30$; other crossover cells use $n=10$; most secondary arms use $n=3$--10 (the scratchpad control also has $n=30$ cells in both protocols). Only the 120-link full-vs-curated comparisons (development grid and final protocol) and the accompanying $n=30$ scratchpad cells are adequately powered; all other results are directional. The scaling of $n$ was itself informative: $n=3$ overstated the 60-link collapse (0/3 versus 6/10).
\item \textbf{Benchmark--system co-evolution, with its measured consequence.} The chain family rewards the statistics the fold tracks, and the requirements were developed against these tasks and seeds. Held-out seeds are consistent so far (3/3 at 60; 2/3 at 120), but the altchain boundary confirms that the advantage is operation-conditional. Task generators authored by others remain the decisive test.
\item \textbf{The curated arm bundles selection with computation.} The control family (scratchpad, masked-history hybrid, calculator, retrieval) brackets the mechanism, and a two-arm decomposition addresses the computation-versus-boundedness question (view+full-history 10/10; view-minus-aggregates 4/10; \S\ref{sec:5.2}). The same decomposition on operations the fold does not precompute remains untested.
\item \textbf{COMPREHEND scope.} The evaluation uses schema-co-designed questions, capped-tail baselines, a size-selected withheld corpus with development exposure (disclosed in Appendix~\ref{app:B}), an LLM reader as an observer proxy, and a tool-call correlation of 99.95\%, which is high but adapter-derived (the \texttt{dangling} and \texttt{last\_error} questions inherit the residual). The synthetic-corpus replication mitigates auditability concerns but not these design limits.
\item \textbf{Single vendor.} The entire stack (worker, readers, extractors, pricing, caching, refusal behavior) comes from one vendor.
\item \textbf{Trust boundaries are not analyzed.} The curator feeds trace-derived content, including tool outputs, back into the worker's context; verbatim validation establishes provenance, not safety. Prompt-injection analysis, provenance-based policy, and secret redaction are unaddressed.
\item \textbf{Single-session traces.} Multi-session and multi-agent ledgers remain untested.
\item \textbf{Retrieval is excluded from chain tasks by construction}, since search would shortcut them; the retrieval-over-own-trace policy partially substitutes, but a task family in which retrieval is available to all arms yet insufficient alone remains needed.
\end{itemize}

\section{Artifacts}\label{sec:8}

We release the tracelab implementation; all four benchmark harnesses; the workbench with all task families and pre-committed schedules; every workbench run trace (synthetic; COMPREHEND panel logs are excluded together with the corpus they embed); the seeded COMPREHEND corpus generator; the scoreboard recording every variant, including failures; the spend ledger; and 99 regression tests. The code, benchmarks, and traces are released at \url{https://github.com/SalesforceAIResearch/tracelab} (BSD-3-Clause); the synthetic corpus is released at \url{https://huggingface.co/datasets/Salesforce/tracelab-comprehend} (CC-BY-4.0). The twelve real transcripts are withheld because they are personal working sessions; the instrument can be rerun on any reader's own transcripts. Benchmark scores are restated from the released scoreboard; CONTINUE results reproduce end-to-end from released code and traces; the synthetic corpus regenerates byte-identically from code, though re-scoring additionally requires the named model endpoints while they remain served. The curated view also registers as an agent inside $\tau^2$-bench retail~\cite{barres2025tau2} via the official agent-factory extension and completed 80 simulations at prompt-token parity with the stock agent; the symmetric baseline lost 27 of 80 simulations to failures in our runner, so no comparative claim is made; the simulation record appears in the released scoreboard, and the integration itself lives in the separate evaluation harness.

\appendix

\section{The development ladder (how the requirements were found)}\label{app:A}

The requirements were not designed in advance; they were extracted from failures during sequential development, with each fix pinned by a regression test before development proceeded.

The deterministic arc (all chain-30 unless noted) unfolded as follows. The v1 structure-only view scored 0.20 on scatter: workers re-read files whose contents the view referenced but did not carry, yielding requirement 1 (content pinning recovered 1.000 on paired seeds). v4 collapsed to 0.000 on chains because the fact store was newest-wins, so thirty same-key deltas became one (requirement 2). v5 remained at 0.000 because the renderer silently kept the last 40 of $\sim$61 facts (requirement 3). v6 exposed value collisions: (key,value) dedup merged legitimately equal deltas from different files (requirement 4; source scoping reached 0.667). The residual failure in v7 was end-stage arithmetic over 30 values (requirement 5; deterministic aggregates reached 5/5 at $\sim$60\% of the full arm's cost). At 120 links, cap eviction silently dropped early values (requirement 6, aggregate-preserving eviction), and a re-read after folding double-counted by exactly +65 (requirement 7).

The extraction arc (prose-chain) followed. Both extractor models fragmented one accumulator into phrasing-dependent keys (contribution/adjustment/advancement), yielding requirement 8: feeding the established key vocabulary forward took the small extractor from 0.667 to 1.000 at unchanged cost. The frontier extractor then returned empty output with refusal stop-reasons on whole batches of individually benign machine noise, $\sim$37/85 calls, yielding requirement 9: bisection with a different-model fallback recovered the score from 0.000 to 0.667. Its residual +41 over-count came from batch-composition nondeterminism; offline replay of the same trace extracted 60/60 exactly (requirement 10, verbatim validation; recovered seeds produce exact totals without raising the success rate). Verbatim validation catches fabricated values but not omissions, mis-keyed sources, or misinterpretation; those are bounded only by the aggregate-vs-oracle recount.

Two later corrections belong to the same pattern. The summarization correction (\S\ref{sec:5.2}) arose because the 400-word cap was our own silent truncation, identified by external review. Requirement 11 closed the same way: external review of the five-failure forensics called for a fix rather than a deferral, and the coverage stamp recovered 5/5 failing seeds while 5/5 passing controls held.

\section{Protocol details}\label{app:B}

\textbf{Seeds.} Development: 1--3. $n=10$ grids: 1--3 plus 4--6, 10--13. $n=30$ extensions: plus 14--33. Held out from the curated/full development ladder and headline grids: 7--9 (later control-arm extensions cover seeds 4--33). Error schedules are pre-committed per (seed, call index) on a dedicated RNG; \texttt{done} is exempt; validation precedes injection, so hallucinated tools receive UnknownTool rather than a retryable error.

\textbf{Workers and readers.} All worker-loop and reader calls use one sample, default decoding, and thinking disabled (the endpoint enables thinking by default, which starved max\_tokens on extraction calls when first encountered; we disable it for all call types and note it as a documented pitfall). Models are named as Sonnet 5 / Haiku 4.5 tiers; the model identifier returned by the API is recorded on every call in the released traces. The fold and the flat log cover main-agent events (both exclude subagent streams); the raw tail is the unfiltered file. COMPREHEND readers are called once per (transcript, condition) with that transcript's questions batched; accuracy is the unweighted mean over per-question scores in [0,1]; 70 of 72 possible questions exist because two transcripts lack a question type's ground truth.

\textbf{Corpus disclosure.} The twelve real transcripts are the authors' own sessions from one framework, selected by size (a criterion that favors long, tool-heavy runs and maximizes the tail-budget gap); the pool overlaps files used for adapter performance testing; question generation and grading were fixed before v1 scoring; and the v0 pilot drew from the same pool. We therefore classify the real-corpus cells as exploratory evidence that outside parties cannot audit; the synthetic replication is the auditable analogue. The synthetic corpus regenerates byte-identically from seeds 201--212, and its generator computes ground truth independently of the adapter (zero mismatches on turns, top tool, files, and latest ask across all twelve sessions). It is distributionally narrower than the real corpus: median 1.1 MB per session vs. 11.0 MB real; 391 vs. 936 events; no thinking, attachment, or session-meta records (so tool calls are 46\% of synthetic events vs. 19\% of real); and error signals differ by construction: injected error-flagged tool results give synthetic sessions a median of 14 per session, whereas organic error-kind records in the real corpus are rare (median 0.5 per session in the released distribution comparison). It replicates the mechanism under test (budget vs. aggregation), not the population of real sessions.

\textbf{Renderer versioning.} Cells keyed v35 and earlier ran the pre-stamp renderer; cells keyed v36 and later ran the shipped post-stamp renderer (run keys are not strictly chronological: the change was made for the v38 coverage-stamp cells, and v36--v37 were executed after it). Every cell's renderer variant is recoverable from its run key; the shipped code contains the post-stamp renderer, and the pre-stamp variant differed only in the header wording and the absence of the coverage stamp described in \S\ref{sec:3}.

\textbf{Independent oracle.} A from-scratch recount that parses raw JSONL directly (never the event pipeline) reproduces every per-key fact and aggregate sum on five chain-120 traces with 0 mismatches (the recount script, \texttt{tools/recount\_oracle.py}, ships with the artifacts). Bookkeeping fields are separately oracle-checked (8/8, mutation-tested). Token accounting: 100K chars of escape-dense JSONL measured 57.8K tokens (1.73 chars/token). View growth on a chain-120 run: 4.7K$\rightarrow$8.0K chars across the final three quarters of the run (one trajectory at one horizon; we make no scaling-rate claim).

\section{Full cell inventory (CONTINUE)}\label{app:C}

\begin{table}[p]\centering\footnotesize\setlength{\tabcolsep}{4pt}
\caption{Full CONTINUE cell inventory. Costs are per attempted run unless noted; altchain rows are the \S\ref{sec:5.3} boundary family.}\label{tab:inventory}
\begin{tabular}{>{\raggedright\arraybackslash}p{4.8cm}ccc>{\raggedright\arraybackslash}p{4.2cm}}
\toprule
arm & 30 & 60 & 120 & \$/run @120 \\
\midrule
full context (flat) & 10/10 & 6/10 & 7/30 & \$7.49 \\
full context (conversational, cached) & --- & 1/3 & 0/10 & $\sim$\$1.06 \\
curated view & 10/10 & 8/10 & 25/30 & \$1.93 (cached 3/3 $\cdot$ \$1.76) \\
scratchpad (flat) & --- & 3/3 & 3/3 & \$6.94 \\
scratchpad (cached) & --- & --- & 26/30 & \$1.04 ($n=3$ pilot: \$1.10) \\
calculator tool (flat) & --- & --- & 10/10 & \$14.88 \\
retrieval-over-trace (last-5 raw steps + top-10 past steps by token-overlap relevance to the current observation, time-reordered) & --- & 0/10 & 0/10 & $\sim$\$1.52 \\
mask (v1-era cell at 30 links excluded) & --- & 0/3 & --- & --- \\
mask + notes hybrid & --- & --- & 10/10 & $\sim$\$2.98 \\
curated + coverage stamp (post-hoc fix cells: 5 failing + 5 passing seeds) & --- & --- & 10/10 & $\sim$\$1.74 \\
curated, shipped post-stamp renderer (full 30-seed rerun, same seeds) & --- & --- & 29/30 & \$1.77 \\
no-injection replication (pilot): curated / full & --- & --- & 5/5 / 0/5 & \$1.60 / $\sim$\$6.83 \\
final-protocol cell: shipped renderer, no injection, $n=30$: curated / full / scratchpad & --- & --- & 30/30 / 8/30 / 30/30 & \$1.59 / \$7.13 / \$0.97 \\
deconfound A: view + full history (post-stamp) & --- & --- & 10/10 & \$7.63 \\
deconfound B: bounded view, aggregates stripped, all facts in view & --- & --- & 4/10 & $\sim$\$2.14 \\
minimal harness accumulator (regex tally + last-5 steps; no trace model) & --- & --- & 10/10 & \$0.67 \\
summary (400-word cap) & 0/3 & 0/3 & not run (stopped at 60) & --- \\
summary (uncapped) & 3/5 & 4/5 & 3/10 & $\sim$\$1.06 @60 $\cdot$ $\sim$\$2.84 @120 \\
altchain: full / curated & --- & 6/10 / 3/10 & 0/10 / 0/10 & $\sim$\$6.00 / $\sim$\$2.49 \\
altchain: scratchpad (cached) & --- & 5/5 & 9/10 & $\sim$\$1.14 \\
\bottomrule
\end{tabular}
\end{table}

We additionally report a heuristic cost per \emph{successful} run at 120 links, computed as mean cost divided by success rate. This metric penalizes inexpensive failure, but it is not an operational forecast, since failures on a fixed seed are not independent retries. Computed from the development-era 120-link cells, this metric gives: full context \$32.1, curated \$2.32, cached scratchpad \$1.20, minimal tally \$0.67, calculator \$14.88, hybrid \$2.98. All other CONTINUE costs in this paper are per attempted run. The chain-60 cache-study diagnostic cells ($n=3$) are reported in \S\ref{sec:5.4} and Appendix~\ref{app:D} rather than duplicated here.

Prose-chain (60 links, $n=3$): full 0.333 $\cdot$ \$2.29; fixed-pattern parser 0.000 $\cdot$ \$0.62; tuned-pattern parser (templates known) 1.000 $\cdot$ \$0.63; small-model parser 1.000 $\cdot$ \$0.80 (extraction \$0.023); frontier parser cascade 0.000$\rightarrow$0.333$\rightarrow$0.667$\rightarrow$0.667 (\$0.066$\rightarrow$\$0.235$\rightarrow$\$0.231$\rightarrow$\$0.238 extraction). The small-model parser at 120 links scored 3/3, with extraction cost \$0.046 and zero refusals.

\section{Cache accounting detail}\label{app:D}

A naive single-block breakpoint (single diagnostic cell) produced 0 cache reads and 512,521 cache-write tokens at the 1.25$\times$ premium, costing \$2.06 versus \$2.03 uncached. At chain-60 ($n=3$/arm), full context fell from \$2.03 (flat, uncached) to \$0.374 (conversational, cached) (5.4$\times$; 501K reads, 41K writes per run), while curated fell only from \$0.74 to \$0.66 (46K reads, 158K writes; refresh flushes dominate). Varying the refresh interval at chain-60 ($n=3$/point) to every 5/10/20 steps yielded \$0.66/\$0.61/\$0.55, with success 1.000 throughout; writes fell 158K$\rightarrow$123K as reads rose 46K$\rightarrow$143K. Curated cached at 120: 3/3, \$1.76. Scratchpad cached at 120 used approximately 2.39M cache-read tokens per run: \$1.10 in the $n=3$ pilot, \$1.04 mean over the $n=30$ grid.

\section{COMPREHEND detail}\label{app:E}

Question types: latest user ask (substring), turn count (exact), most-used tool (exact), files touched (set-F1), last error (substring), dangling calls (set-F1-or-none). Per-question scores on the real corpus (view condition, Sonnet 5): latest\_ask 1.0, top\_tool 1.0, files 1.0, dangling 0.917; n\_turns and last\_error account for the remaining error. The files question scores 0.126 in the raw-tail condition (Sonnet 5) --- aggregation questions are unanswerable from a capped tail of a large transcript --- and the turn-count question scores 0.0 even in the flat-log condition; both are consequences of the retention mechanism rather than of reader failure. The dangling question scores 0.917 in every condition because eleven of twelve transcripts share the majority ``none'' answer; a constant ``none'' responder matches every condition on this item. Macro-averaging over the remaining five question types moves the conditions apart rather than together: view 0.850 / raw 0.385 / flat 0.554 (Sonnet 5) and 0.815 / 0.382 / 0.432 (Haiku 4.5), so the degenerate item understates the condition effect. A v0 pilot (3 same-pool transcripts, Sonnet 5) gave view 0.889 versus raw 0.347. Synthetic corpus (12 sessions, 11.4 MB): view 1.000/0.944, flat 0.722/0.638, raw 0.673/0.600 (Sonnet 5/Haiku 4.5).

\bibliographystyle{unsrtnat}
\bibliography{main}
\end{document}